\documentclass[default,iicol]{sn-jnl}

\jyear{2023}%

\theoremstyle{thmstyleone}%
\theoremstyle{thmstyletwo}%

\theoremstyle{thmstylethree}%

\usepackage{graphicx}
\usepackage{subfigure}
\usepackage{tabularx}
\usepackage{multirow}
\usepackage{wrapfig}
\usepackage{color,colortbl}
\definecolor{Gray}{gray}{0.9}
\usepackage{amsmath}
\usepackage{amssymb}
\usepackage{tabularx}
\usepackage{multirow}
\usepackage{multicol}
\usepackage{booktabs}
\usepackage{bbm}
\usepackage[T1]{fontenc}
\usepackage{multirow}
\usepackage{multicol}

\newcolumntype{L}[1]{>{\raggedright\arraybackslash}m{#1}}
\newcolumntype{C}[1]{>{\centering\arraybackslash}p{#1}}

\begin{document}

\title[Article Title]{DIVOTrack: A Novel Dataset and Baseline Method for Cross-View Multi-Object Tracking in \textbf{DIV}erse \textbf{O}pen Scenes}


\author[1]{\fnm{Shengyu} \sur{Hao}}\email{shengyuhao@zju.edu.cn}
\equalcont{These authors contributed equally to this work.}

\author[1]{\fnm{Peiyuan} \sur{Liu}}\email{peiyuan.19@intl.zju.edu.cn}
\equalcont{These authors contributed equally to this work.}

\author[2]{\fnm{Yibing} \sur{Zhan}}\email{zhanyibing@jd.com}

\author[1]{\fnm{Kaixun} \sur{Jin}}\email{3710111702@zju.edu.cn}

\author[1]{\fnm{Zuozhu} \sur{Liu}}\email{zuozuliu@intl.zju.edu.cn}

\author[3]{\fnm{Mingli} \sur{Song}}\email{brooksong@zju.edu.cn}

\author[4]{\fnm{Jenq-Neng} \sur{Hwang}}\email{hwang@uw.edu}

\author*[1]{\fnm{Gaoang} \sur{Wang}}\email{gaoangwang@intl.zju.edu.cn}


\affil[1]{\orgdiv{ZJU-UIUC Institute}, \orgname{Zhejiang University}, \orgaddress{\city{Haining}, \country{China}}}

\affil[2]{\orgdiv{JD Explore Academy}, \orgaddress{\city{Beijing}, \country{China}}}

\affil[3]{\orgdiv{College of Computer Science and Technology}, \orgname{Zhejiang University}, \orgaddress{\city{Hangzhou}, \country{China}}}

\affil[4]{\orgdiv{Department of Electrical \& Computer Engineering}, \orgname{University of Washington}, \orgaddress{\city{Seattle}, \country{USA}}}

\abstract{Cross-view multi-object tracking aims to link objects between frames and camera views with substantial overlaps.
Although cross-view multi-object tracking has received increased attention in recent years, existing datasets still have several issues, including 1) missing real-world scenarios, 2) lacking diverse scenes, 3) containing a limited number of tracks, 4) comprising only static cameras, and 5) lacking standard benchmarks, which hinder the investigation and comparison of cross-view tracking methods. 
To solve the aforementioned issues, we introduce \textit{DIVOTrack}: a new cross-view multi-object tracking dataset for \textbf{DIV}erse \textbf{O}pen scenes with dense tracking pedestrians in realistic and non-experimental environments. Our DIVOTrack has fifteen distinct scenarios and 953 cross-view tracks, surpassing all cross-view multi-object tracking datasets currently available.
Furthermore, we provide a novel baseline cross-view tracking method with a unified joint detection and cross-view tracking framework named \textit{CrossMOT}, which learns object detection, single-view association, and cross-view matching with an \textit{all-in-one} embedding model. Finally, we present a summary of current methodologies and a set of standard benchmarks with our DIVOTrack to provide a fair comparison and conduct a comprehensive analysis of current approaches and our proposed CrossMOT. The dataset and code are available at \url{https://github.com/shengyuhao/DIVOTrack}.}




\maketitle

\section{Introduction}\label{sec1}

Single-view multi-object tracking (MOT) has been extensively explored in recent years \cite{tracktor_2019_ICCV,wang2020towards,centertrack,wang2019exploit,zhang2021fairmot,wang2022recent,braso2022multi}. However, the limitation of the single viewpoint causes occluded objects to be lost in long-term tracking \cite{wang2021track,wang2022split}.
The above issue can be alleviated under tracking with the cross-view setting
\cite{hofmann2013hypergraphs,xu2017cross,han2020cvmht}. 
Specifically, given multiple synchronized videos capturing the same scene from different viewpoints, there is a high probability that an object obscured in one view is visible in another. Cross-view settings can compensate for the occlusion information of single-view monitoring with their complementary information.
Due to its effectiveness, cross-view multi-object tracking has attracted considerable interest, and numerous cross-view tracking methods \cite{ayazoglu2011dynamic,hofmann2013hypergraphs,tang2018joint,fleuret2007multicamera,liu2016multi,xu2017cross,han2020cvmht,gan2021mvmhat} have been proposed in the literature. For example, some cross-view tracking methods focus on excavating information from multi-views \cite{ayazoglu2011dynamic,hofmann2013hypergraphs,tang2018joint}. Some methods explore new formulations and solutions to the problem \cite{fleuret2007multicamera,liu2016multi}. Moreover, some recent works \cite{xu2017cross,han2020cvmht} apply graph clustering to the cross-view tracking problem. 

However, due to the limitations of current cross-view tracking datasets, several significant challenges still exist when comparing current approaches for cross-view tracking and exploring new ones. On the one hand, although various cross-view tracking datasets \cite{fleuret2007multicamera,xu2017cross,chavdarova2018wildtrack,gan2021mvmhat} have appeared in recent years, these existing datasets have significant drawbacks. 
To be specific, existing datasets suffer from 1) missing real-world scenarios, 2) lacking diverse tracking scenes, and 3) containing a limited number of tracks. Hence, the datasets can be hardly used to test the efficacy of cross-view tracking approaches comprehensively. Moreover, the vast majority of videos in known datasets were captured with static cameras, hampering research on tracking algorithms for moving cameras.


To overcome the aforementioned difficulties and facilitate future research on cross-view tracking, we present a novel cross-view dataset for multi-object tracking in \textbf{DIV}erse \textbf{O}pen Scenes, dubbed \textit{DIVOTrack}.
In particular, our DIVOTrack dataset has the following primary characteristics:
1) DIVOTrack video recordings are captured in real-world circumstances and contain a mixture of a limited number of pre-selected actors and a large number of unwitting participants.
2) DIVOTrack offers diverse scenes. It contains outdoor and indoor scenes with various surrounding environments, such as streets, shopping malls, buildings, squares, and public infrastructures.
3) DIVOTrack provides a large collection of IDs and tracks focusing on crowded settings. It has a total of 1,690 single-view tracks and 953 cross-view tracks, both of which are significantly larger than the previous cross-view multi-object tracking datasets. 
4) DIVOTrack was taken from widely moving cameras, enabling the study of cross-view tracking with moving cameras in the community.

In addition to the proposed DIVOTrack dataset, we propose an end-to-end cross-view multi-object tracking baseline framework named \textit{CrossMOT} to learn object embeddings from multiple views, extended from the single-view tracker FairMOT \cite{zhang2021fairmot}. CrossMOT is a unified joint detection and cross-view tracking framework, which uses an integrated embedding model for object detection, single-view tracking, and cross-view tracking. 
Specifically, CrossMOT uses decoupled multi-head embedding that can learn object detection, single-view feature embedding, and cross-view feature embedding simultaneously. 
To address the ID conflict problem between cross-view and single-view embeddings, we use locality-aware and conflict-free loss to improve the embedding performance. During the inference stage, the model takes advantage of the joint detector as well as separate embeddings for cross-frame association and cross-view matching.

Our main contributions are summarized as follows. 
\begin{itemize}
\item A novel cross-view multi-object tracking dataset is proposed, which is more realistic and diverse, has more crowded tracks, and incorporates moving cameras. The dataset is with high image quality and clean ground truth labels.
\item We propose a novel cross-view tracker termed  \textit{CrossMOT}, which is the first work that extends the joint detection and embedding from the single-view tracker to the cross-view. Our proposed CrossMOT is an all-in-one embedding model that simultaneously learns object detection, single-view, and cross-view features.
\item 
We build a standardized benchmark for cross-view tracking evaluation. Extensive experiments are conducted using baseline tracking methods, including single-view and cross-view tracking. We show that the proposed CrossMOT achieves high cross-view tracking accuracy and significantly outperforms state-of-the-art (SOTA) methods on DIVOTrack, MvMHAT \cite{gan2021mvmhat} and CAMPUS \cite{xu2016multi}. The experiment results can be used as a reference for future research. 
\end{itemize}



The outline of the paper is as follows: In Section~\ref{sec:related_work}, we review state-of-the-art (SOTA) cross-view MOT methods and datasets. Section~\ref{sec:wild_scene} describes the details of the proposed DIVOTrack dataset. We introduce our proposed CrossMOT in Section~\ref{sec:method}. We provide the experiments of baseline methods on the benchmark in Section~\ref{sec:exp}, followed by the conclusion and future work in Section~\ref{sec:conclude}.

\section{Related Work}
\label{sec:related_work}

\subsection{Inter-Camera Tracking}
In some existing methods, inter-camera tracking \cite{tesfaye2017multi,cai2014exploring,lee2017online,tang2018single,hsu2019multi,hsu2021multi,ma2021deep} does not assume overlapping views between cameras. Usually, an object may leave the view of one camera and then enter the view of another camera. Research in this category attempts to match single-camera trajectories across non-overlapping cameras by exploiting intrinsic information of objects, such as appearance features \cite{tesfaye2017multi,cai2014exploring}, motion patterns \cite{hofmann2013hypergraphs}, and camera topological configuration \cite{lee2017online}. For appearance cues, \cite{zhang2017multi} uses convolutional neural networks (CNNs) to generate the feature
representation for each target and proposes a feature re-ranking mechanism to find correspondences among tracklets. \cite{ristani2018features} considers not only the CNN-based appearance features but also motion patterns. Moreover, it formulates the inter-camera MOT task as a binary integer program problem and proposes the deep feature correlation clustering approach to match the trajectories of a single camera to all other cameras. Some works consider the camera topology in inter-camera MOT. For example, \cite{cheng2017part} attempts to match local tracklets between every two neighboring cameras.

\subsection{Cross-View Tracking}
Cross-view tracking is one specific category of inter-camera tracking with shared large overlapping views among different cameras. Cross-view tracking has not been widely explored due to the challenges of data collection, cross-view object association, and multi-modality feature fusion. 
Some existing methods focus on excavating multi-view information based on overlapping views, such as \cite{ayazoglu2011dynamic,khan2001human,hofmann2013hypergraphs,tang2018joint}. Some focus on new problem formulations and solutions, such as \cite{fleuret2007multicamera,liu2016multi}.
Recent works \cite{xu2017cross,han2020cvmht} formulate cross-view tracking as a graph clustering problem. The graph is constructed with detections or tracklets as nodes. Afterward, the similarities between nodes are measured with appearance and motion features. However, the similarity measure is based on hand-crafted feature fusion, which may be sub-optimal. Besides, optimization for graph clustering in the inference stage is usually computationally expensive. How to automatically combine features from different modalities is still an open question in the cross-view tracking area.

\begin{figure*}[!ht]
\begin{center}
\includegraphics[width=0.9\linewidth]{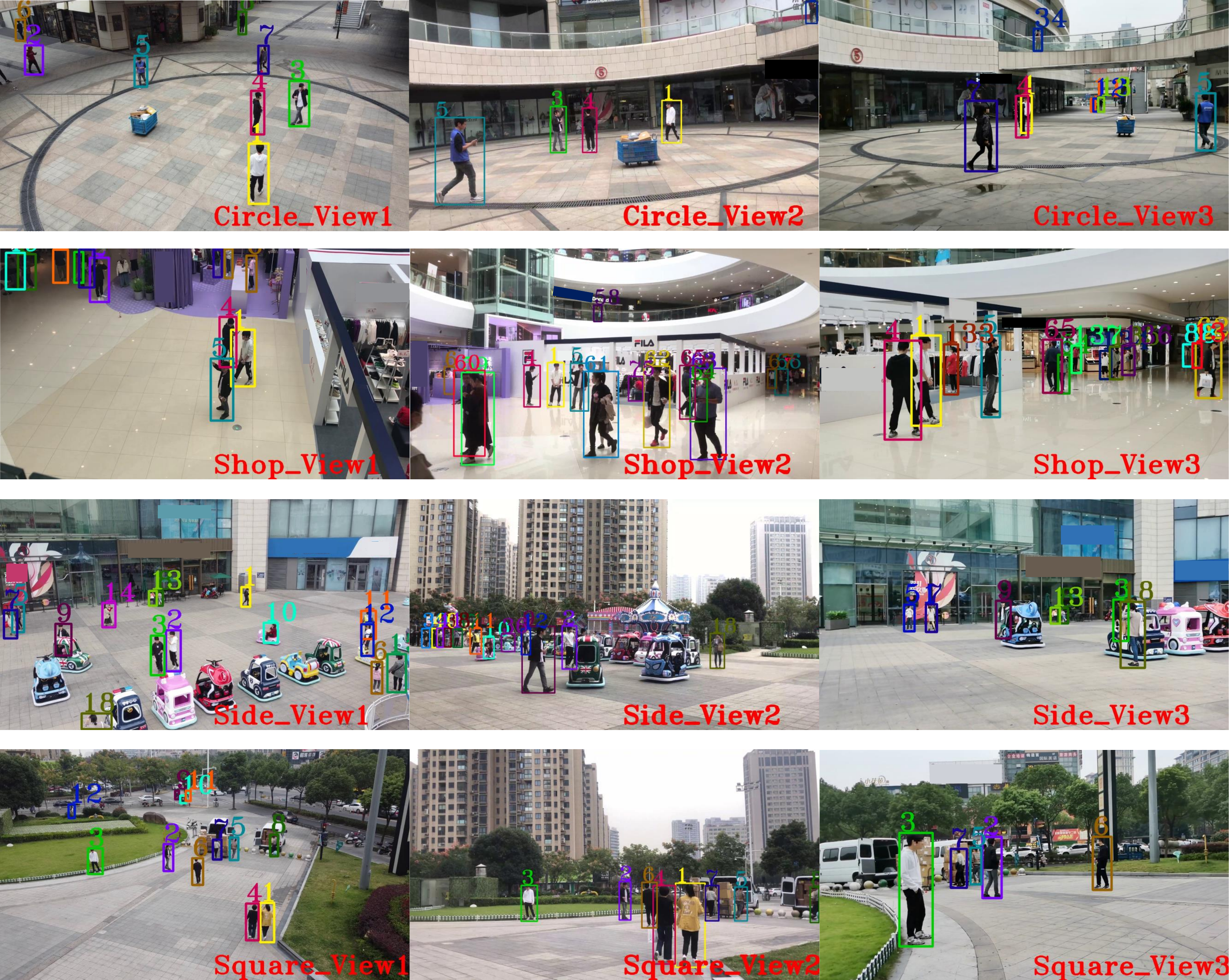}
\end{center}
   \caption{Examples of the DIVOTrack dataset. From top to bottom: \textit{Circle}, \textit{Shop}, \textit{Side}, and \textit{Ground} scenes of three views, respectively. The same person that appears in different views is shown in the same color.}
\label{fig:example}
\end{figure*}

\subsection{Cross-View Tracking Datasets}

There are several existing commonly used cross-view multi-object tracking datasets, including EPFL \cite{fleuret2007multicamera}, CAMPUS \cite{xu2016multi}, WILDTRACK \cite{chavdarova2018wildtrack}, and MvMHAT \cite{gan2021mvmhat}. EPFL dataset is one of the traditional cross-view tracking datasets, captured in three or four different views by static cameras. The major limitation of this dataset is that almost all sequences are captured in a controlled setting, which is far from real-world scenarios. Besides, the videos have very low resolutions, causing difficulty in learning informative appearance embeddings of the objects. CAMPUS dataset contains more realistic scenarios. However, most subjects are pre-selected, and the ground truth annotations are not very accurate. WILDTRACK is captured in an outdoor square with crowded pedestrians. However, it only contains one single scene and some of the pedestrians are not annotated, hindering the usage of the dataset. MvMHAT is one of the recently released datasets. MvMHAT still suffers from a very limited number of subjects, and all the video recordings are collected in an identical and experimental environment. Compared with these datasets, our DIVOTrack is more realistic and diverse, has more crowded tracks, and incorporates dynamic scenes.

\begin{figure*}
\centering
\includegraphics[width=1\linewidth]{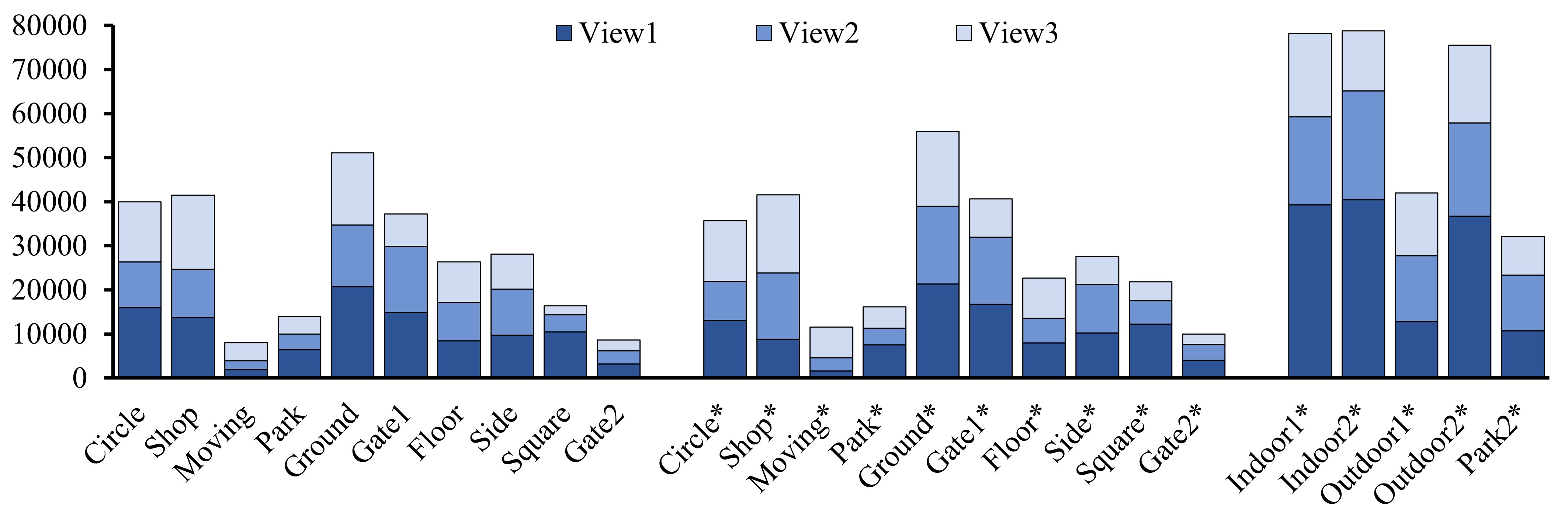}
\caption{Numbers of boxes for each view on training and test set, respectively. The different colors of each bar represent different views.``*'' represents the test set. View1 is the drone, and View2 and View3 are the cell phones. For View2 of the first ten scenes, the train set and test set are captured by the higher-res cell phone.}
\label{fig:box}
\end{figure*}
\section{DIVOTrack Dataset}
\label{sec:wild_scene}

We present a self-collected cross-view dataset for diverse open scenes, namely DIVOTrack, to facilitate cross-view tracking research in the community. We explain the dataset collection, annotation, and statistics as follows.

\subsection{Data Collection}
We collect data in 15 different real-world scenarios, including indoor and outdoor public scenes. We capture all the sequences by using three moving cameras and are manually synchronized. We pre-select a specific overlapping area among different views for each scene. We require all the cameras to shoot at the pre-selected area with random smooth movements. 

We collect the data from fifteen diverse open scenes with varying population densities and public spaces. And our dataset contains twelve outdoor scenes and three indoor scenes from streets, shopping malls, gardens, and squares, namely \textit{Circle}, \textit{Shop}, \textit{Moving}, \textit{Park}, \textit{Ground}, \textit{Gate1}, \textit{Floor}, \textit{Side}, \textit{Square}, \textit{Gate2}, \textit{Indoor1}, \textit{Indoor2}, \textit{Outdoor1}, \textit{Outdoor2} and \textit{Park2}. There are both moving dense crowds and sparse pedestrians in outdoor and indoor scenes. The surrounding environment of outdoor scenes is diverse, including streets, vehicles, buildings, and public infrastructures. Meanwhile, the indoor scene comes from a large shopping mall, with a more complicated and severe occlusion of the crowd than the outdoor environment. 

We record the first ten scenes with three types of moving cameras: one is mounted on a flying UAV with a resolution of $1920\times 1080$, overlooking the ground with a pitch angle of around $45^{\circ}$; the other two cameras are from mobile phones held by two people, with the resolution of $3640\times 2048$ and $1920\times 1080$, respectively. We collected the remaining five scenes using the same devices, all of which were recorded at a resolution of $1920\times 1080$. All the raw video recordings are with 60 FPS. We record each scene using all three cameras and divide the 15 scenes into three parts. For the scenes \textit{Circle}, \textit{Shop}, \textit{Moving}, \textit{Park}, \textit{Ground}, \textit{Gate1}, \textit{Floor}, \textit{Side}, \textit{Square} and \textit{Gate2}, we split each scene into two sets: one for the train set and the other for the easy test set. For the remaining scenes \textit{Indoor1}, \textit{Indoor2}, \textit{Outdoor1}, \textit{Outdoor2} and \textit{Park2}, we select them as the hard test set to assess the generalization of the trackers.
It should be noted that during the recording process, both the UAV and the mobile phone camera have a certain degree of shaking, which is normal for moving camera-based recording.

After recording, we synchronize the videos manually. We align the timestamps with the beginning and ending frames of each recording batch. Since the FPS for all cameras is 60, the synchronization error ranges between -1/120 and +1/120 milliseconds (ms), which is bounded by 8 ms. Because pedestrian movement is not fast, the synchronization error for the pedestrian tracking task is acceptable. After alignment, each video is downsampled to 30 FPS. For people who are close to the camera, we also add mosaics on human faces.

\begin{figure*}
\centering
\includegraphics[width=1\textwidth]{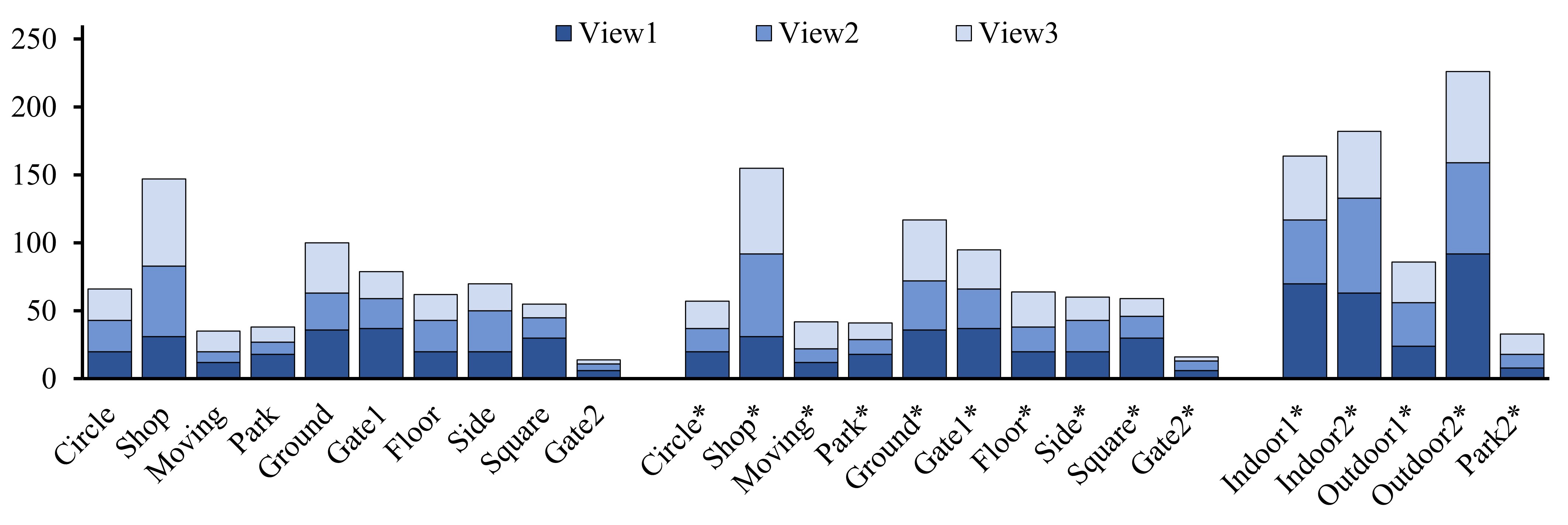}

\caption{Number of tracks for each view on training and test set, respectively. The different colors of each bar represent different views. ``*'' represents the test set. View1 is the drone, View2 and View3 are the cell phones. For View2 of the first ten scenes, the train set and test set are captured by the higher-res cell phone.} \label{fig:track}
\end{figure*}

\begin{figure*}
\centering
\includegraphics[width=1\textwidth]{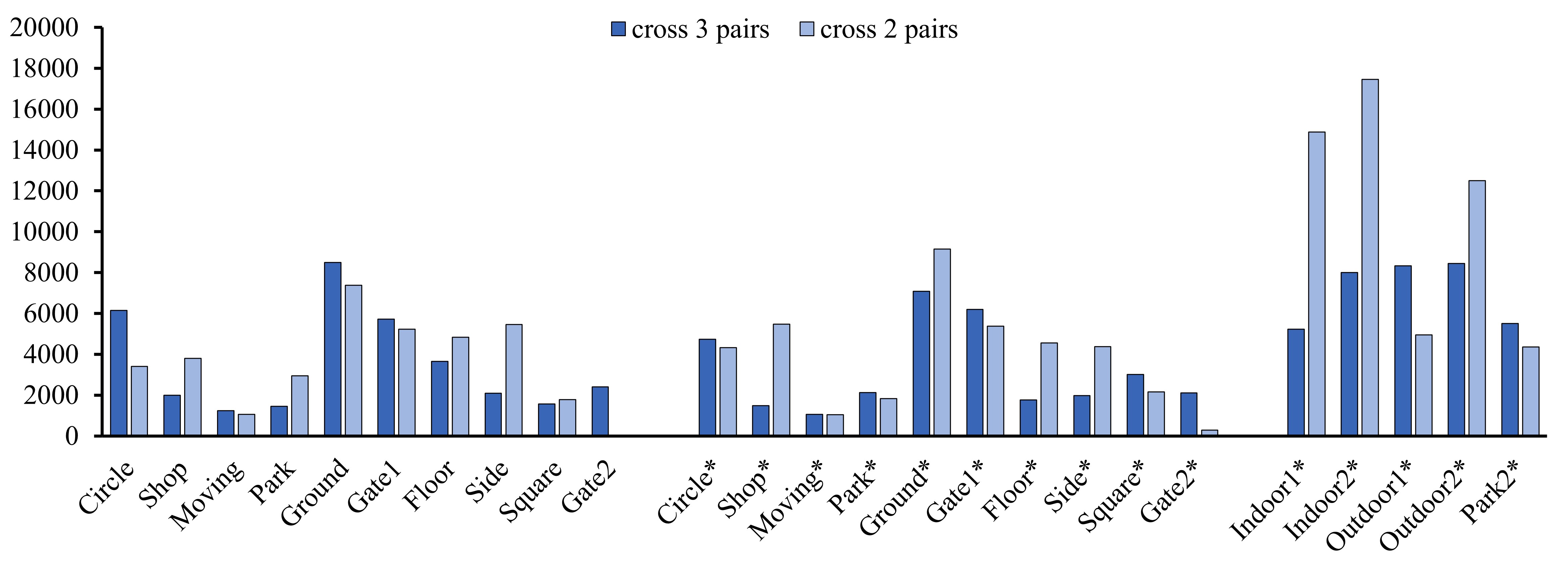}

\caption{Number of cross-view boxes for each scene on the train and test set, respectively. ``*'' represents the test set. View1 is the drone, and View2 and View3 are the cell phones. View2 of the train set and test set of the first ten scenes is the high-res cell phone.} \label{fig:cv_crossbox}
\end{figure*}

\subsection{Data Annotation}

For data annotation, we aim to obtain the full-figure ground-truth bounding box and global pedestrian ID across different views in each scene. We bound the full extent of the person's body rather than the visible part of it. The data annotation contains three main steps, \textit{i.e.}, track initialization, single-view correction, and cross-view matching. The annotation process is demonstrated as follows.

We utilize a pre-trained single-view tracker to initialize the object bounding boxes and tracklets, which can significantly reduce the labor cost of annotation. Specifically, we use CenterTrack \cite{centertrack} to generate the raw tracks with the model pre-trained on the MOT Challenge dataset \cite{milan2016mot16}.

To further save labeling time, we manually correct the tracking results, including both bounding boxes and IDs, for every ten frames. 
After correction, we interpolate the boxes linearly for intermediate frames. Throughout the annotation process, interpolation can result in the presence of amodal boxes, which represent objects that may be partially (but not fully) occluded \cite{khurana2021detecting, reddy2019occlusion}. 
Furthermore, the ground truth generated by CenterTrack \cite{centertrack} introduces bias by naturally predicting similar (or same) boxes that were used to generate the ground truth. This grants an unfair advantage to this approach. To address this, we only labeled the visible objects (those with more than 30\% visibility) and conducted a thorough review of the entire dataset after the interpolation stage.

After the single-view correction, the objects that appear in multiple views are still not matched. We will assign the same global IDs to these identical objects across all views.
Based on the corrected single-view tracklets,  we re-assigned objects that appear in two or three views with the same IDs. We renumber the IDs according to the first time the object appears in any of the three views.
Ultimately, we assign tracklets matched in different views with an identical global ID. And we also assign the tracklet that only appears in a single view with a global ID.

\subsection{Dataset Statistics}
We count the number of bounding boxes and tracks in the train and test sets for each scene. We also show the important statistics of the dataset for both single-view and cross-view tracking.





\subsubsection{Boxes and Tracks}
The detailed bounding box statistics of the DIVOTrack dataset are shown in Fig.~\ref{fig:box}. Our whole DIVOTrack dataset has 830K boxes, of which 270K boxes belong to the train set, 280K belong to the easy test set, and the rest belongs to the hard test set. The different colors of each bar represent different views. The number of bounding boxes reflects the density of crowds in each scene. For example, there are less than 10K boxes in the \textit{Moving} scene but more than 50K boxes in the \textit{Ground} scene, demonstrating a diverse density of the dataset.

The number of tracks is shown in Fig.~\ref{fig:track}. We count the number of tracks from 75 videos. We can observe a large variation in the number of tracks in different scenes. For example, more than 140 tracks in \textit{Shop} but less than 20 tracks in \textit{Gate2} from the train set. Besides that, the proportion of tracks among the three views is very close. 
\begin{figure*}[!t]
\begin{center}
\includegraphics[width=1\linewidth]{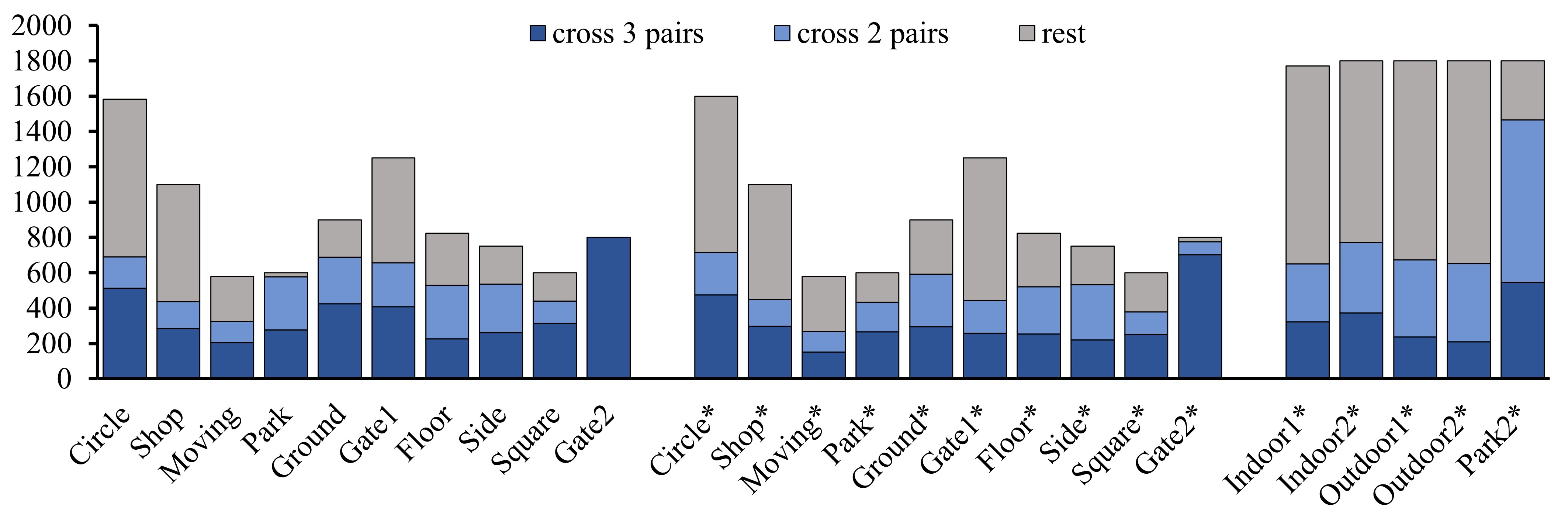}
\end{center}
   \caption{The track duration for each scene on the train and test set, respectively. ``*'' represents the test set. The ordinate axis is the total number of frames and the grey part plus the colored part is the total number of frames.}
\label{fig:cv_duration}
\end{figure*}

\subsubsection{Cross-View Statistics}
To show the cross-view statistics, we plot the number of boxes from the same object across two and three views in Fig.~\ref{fig:cv_crossbox}. There are more boxes across two views than three views in several scenes, such as \textit{Shop}, \textit{Floor}, and \textit{Side}, showing that some pedestrians are not visible from at least one view. This demonstrates that a large view angle variance exists in the dataset. To better present the dataset, we show some sampled frames of the dataset in Fig.~\ref{fig:example}. From top to bottom are examples of \textit{Circle}, \textit{Shop}, \textit{Side}, and \textit{Ground} scenes in three views, respectively. The same person that appears in different views is shown in the same color.
We also count the duration of object trajectories appearing across multiple views, as shown in Fig.~\ref{fig:cv_duration}. The colored bars for each scene represent the average duration of pedestrian appearance across different views. The cross-view overlapping duration accounts for over half of the total time, demonstrating that our dataset has sufficient cross-view tracklets.



\begin{table*}[!t]
\small
\caption{Comparison between cross-view multi-object tracking datasets.}
\centering
\begin{tabular}{l|ccccc}
\toprule
Attribute & EPFL  & CAMPUS & MvMHAT & WILDTRACK & \textbf{DIVOTrack} \\
\midrule
Scenes & 5 & 4 & 1 & 1 & \textbf{15}\\
Groups & 5 & 4 & 12 & 1 & \textbf{25}\\
Views & 3-4 & 4 & 3-4 & \textbf{7} & 3\\
Sequences & 19 & 16 & 46 & 7 & \textbf{75}\\
Frames & \textbf{97K} & 83K & 31K & 3K & 81K\\
Single-View Tracks & 154 & 258 & 178 & - & \textbf{1,690}\\
Cross-View Tracks & 41 & 70 & 60 & 313 & \textbf{953}\\
Boxes & 625K & 490K & 208K & 40K & \textbf{830K}\\
Moving Camera & No & No & \textbf{Yes} & No & \textbf{Yes}\\
Subject & Actor & Actor & Actor & \textbf{Mixed} & \textbf{Mixed}\\

\bottomrule
\end{tabular}
\label{tab:dataset}
\end{table*}

\subsection{Comparison with Existing Datasets}

We compare several existing cross-view multi-object tracking datasets with our DIVOTrack, namely EPFL \cite{fleuret2007multicamera}, CAMPUS \cite{xu2016multi}, MvMHAT \cite{gan2021mvmhat}, and WILDTRACK \cite{chavdarova2018wildtrack}.
Most existing datasets have non-diverse scenes and a limited number of tracking objects. 
Specifically, the EPFL dataset \cite{fleuret2007multicamera} contains five sequences: Terrace, Passageway, Laboratory, Campus, and Basketball. In general, each sequence consists of three or four different views and films with 6-11 pedestrians walking or running around, lasting 3.5-6 minutes. Each view is shot at 25 FPS with a relatively low resolution of $360\times 288$.
CAMPUS dataset \cite{xu2017cross} contains four sequences, \textit{i.e.}, two gardens, one parking lot, and one auditorium, shot by four 1080P cameras. The recorded videos last three to four minutes with 30 FPS. 
The MvMHAT dataset \cite{gan2021mvmhat} contains 12 video groups and 46 sequences, where each group includes three to four views. The videos are collected with four wearable cameras, \textit{i.e.}, GoPro, covering an overlapped area with multiple people from significantly different directions, \textit{e.g.}, near 90-degree view-angle difference. The videos are manually synchronized and annotated with bounding boxes and IDs on 30,900 frames.
The WILDTRACK dataset \cite{chavdarova2018wildtrack} is captured by seven static cameras with 60 FPS in seven distinct views. WILDTRACK provides a joint calibration and synchronization of sequences. There are about 3000 annotated frames, 40,000 bounding boxes, and over 300 individuals.

We report the detailed comparison in Table~\ref{tab:dataset}. 
We can observe that the DIVOTrack dataset has four main advantages. 
1) DIVOTrack contains a mixture of a small number of pre-selected subjects and a large number of non-experimental walking pedestrians in the video recording, which is captured in real-world scenarios and is much more realistic than existing datasets. 
2) DIVOTrack has more diverse scenes. 
3) DIVOTrack has a much larger set of IDs and tracks, focusing on more crowded scenarios. 
4) Our dataset was taken from widely moving cameras, enabling cross-view tracking research with moving cameras in the community.

\section{CrossMOT}
\label{sec:method}

To demonstrate the effectiveness of the proposed DIVOTrack dataset and deal with the challenges of cross-view tracking, a baseline cross-view tracking method is highly needed. 

\begin{figure*}[!t]
\centering
\includegraphics[width=\textwidth]{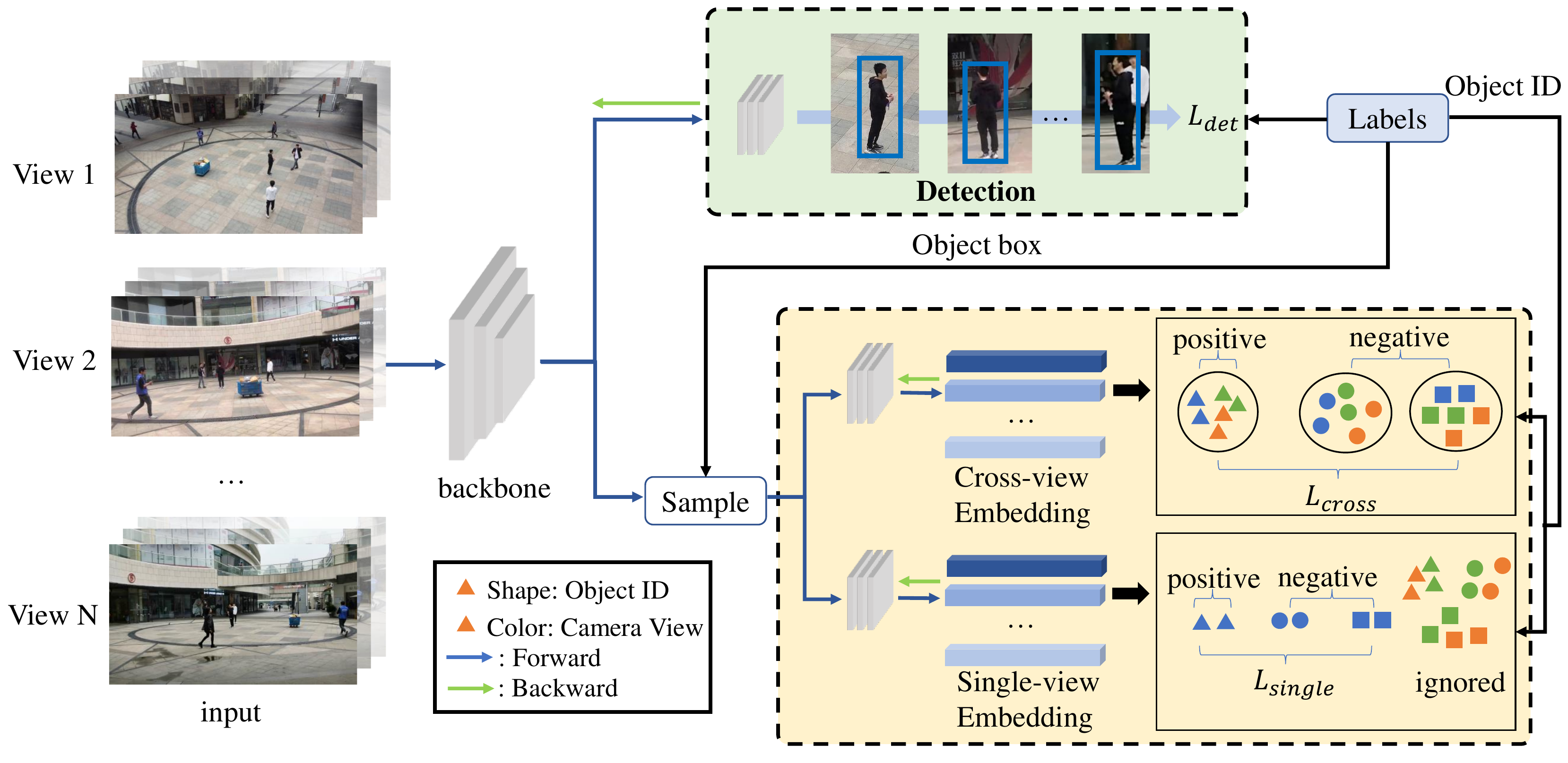} 
\caption{The proposed CrossMOT framework. The input cross-view video clips are fed into the backbone and then followed by three heads for embedding. The blue and green arrows represent the forward and backward flow, respectively. In the diagram at the bottom right, the color of the symbols corresponds to the camera view and their shape corresponds to the global ID.
The detection branch provides the detection results and the other two embedding branches help the single-view and cross-view association. 
}
\label{fig_fw}
\end{figure*}

In this paper, we propose a unified joint detection and cross-view tracking framework, which learns object detection, single-view association, and cross-view matching with an \textit{all-in-one} embedding model as shown in Fig~\ref{fig_fw}.

Our proposed method adopts the decoupled multi-head embeddings that simultaneously learn object detection, single-view re-identification (Re-ID), and cross-view Re-ID. To address the ID conflict issue for cross-view and single-view embeddings, we employ a locality-aware and conflict-free loss to improve the joint embedding.
Specifically, single-view embeddings focus on learning temporal continuity, while cross-view embeddings focus on learning the invariant appearance of objects. As a result, cross-view tracking methods should consider both single-view and cross-view features for embedding. We use two different Re-ID losses for training the single-view and cross-view embeddings. During the inference stage, we employ the single embedding to calculate the similarity among objects within the same view, while the cross-view embedding is utilized to determine the similarity between objects across different views. Ultimately, we combine two embeddings with detection boxes for cross-view multi-object tracking.


In this section, we overview cross-view tracking and our proposed baseline in Sub-section~\ref{sec:overview}. We describe the details of the baseline in Sub-section~\ref{sec:decouple_emb} and provide the inference stage in Sub-section~\ref{sec:inference}.

\subsection{Overview}
\label{sec:overview}
We first demonstrate the cross-view tracking task. Given a set of synchronized video sequences $\mathcal{V}=\{\boldsymbol{V}_1,\boldsymbol{V}_2,...,\boldsymbol{V}_{N}\}$ from multiple views of the same scene, we aim to simultaneously detect and track objects across different views, where $N$ is the number of different camera views, and each video $\boldsymbol{V}_i$ contains $T_i$ successive frames $\{\boldsymbol{I}_1,\boldsymbol{I}_2,...,\boldsymbol{I}_{T_i}\}$. We aim to detect and track objects across multiple views simultaneously, \textit{i.e.}, distinguish identical objects with a shared global ID across frames and views.

We propose a novel cross-view tracker, namely \textit{CrossMOT}. Our proposed CrossMOT adopts the backbone of CenterNet \cite{zhou2019objects}, denoted as $f(\cdot;\boldsymbol{\theta}_f)$, followed by three sub-networks, including a detection head $h_{d}(\cdot;\boldsymbol{\theta}_{d})$, a cross-view Re-ID head $h_{c}(\cdot;\boldsymbol{\theta}_{c})$, and a single-view Re-ID head $h_{s}(\cdot;\boldsymbol{\theta}_{s})$, where $\boldsymbol{\theta}_f$, $\boldsymbol{\theta}_{d}$, $\boldsymbol{\theta}_{c}$, $\boldsymbol{\theta}_{s}$ are model parameters. 
However, when using multiple heads for separate embeddings, the definition of ground truth IDs between single-view and cross-view is different, \textit{i.e.}, the same objects across different videos are regarded as different objects in the single-view tracking since the temporal continuity is disobeyed, which causes conflicts during training.
In the following subsections, we introduce our proposed CrossMOT and illustrate how to decouple multi-head embedding and address the conflict issue in multi-task learning. We also summarize our association method for inference with separate embeddings.

\subsection{Decoupled Multi-Head Embedding}
\label{sec:decouple_emb}
Our framework of CrossMOT is shown in Fig.~\ref{fig_fw}.
The CrossMOT conducts object detection, single-view tracking, and cross-view tracking simultaneously. To fulfill multiple tasks, we decouple the embedding into three head networks: object detection head, cross-view Re-ID head, and single-view Re-ID head. The details are as follows.

\subsubsection{Object Detection Embedding}

The object detection head of our framework follows CenterNet \cite{zhou2019objects}, which includes the prediction of object confidence heatmap $\boldsymbol{h}_{hm}$, object size $\boldsymbol{h}_{size}$, and the object offset $\boldsymbol{h}_{offset}$. The loss is defined as follows,
\begin{equation}
    \mathcal{L}_{d} = \sum_{\boldsymbol{I} \in \mathcal{V}} \boldsymbol{w}_d^T \phi_{d}(h_d(f(\boldsymbol{I};\boldsymbol{\theta}_{f});\boldsymbol{\theta}_{d}), \boldsymbol{y}_{d}),
    \label{loss_det}
\end{equation}
where $\boldsymbol{y}_d$ is the ground truth of object class, size, and location heatmap; $\phi_{d}(\cdot, \boldsymbol{y})$ contains individual losses of the detection, including the focal loss for classification and $l_1$ loss for size and offset regression; $h_{d}(\cdot;\boldsymbol{\theta}_{d}) = \{{\boldsymbol{h}_{hm}},{\boldsymbol{h}_{size}},{\boldsymbol{h}_{offset}}\}$; and $\boldsymbol{w}_d=\{w_{{hm}},w_{{size}},w_{{offset}}\}$ are the weights of individual losses.

\subsubsection{Cross-view Re-ID Embedding}
We employ the cross-view Re-ID embedding to provide the cross-view feature which is used to associate the same person from different views in the same scene. 
First, we extract the cross-view Re-ID embedding of each object based on its center pixel location. 
 
As long as the objects across different views are from the same object, we assign a unique global ID  to the object. The global IDs are unique IDs in the entire train set, including multiple scenes with different views. We follow the conventional cross-entropy loss for the cross-view Re-ID, \textit{i.e.},
\begin{equation}
    \mathcal{L}_{c} = \sum_{\boldsymbol{I} \in \mathcal{V}}  \phi_c(h_{c}(f(\boldsymbol{I};\boldsymbol{\theta}_f);\boldsymbol{\theta}_{h_c}), \boldsymbol{y}^{GID}),
    \label{loss_cross}
\end{equation}
where $\phi_c(\cdot, \cdot)$ is cross-entropy loss, and $\boldsymbol{y}^{GID}$ is the one-hot vector of object global ID. 

\subsubsection{Locality-aware and Conflict-free Single-view Embedding}

With the combination of object detection embedding and cross-view Re-ID embedding, our tracker can already achieve the goal of end-to-end cross-view tracking. However, from our experimental observations, the shared embedding from cross-view Re-ID has degradation on both single-view association and cross-view matching tasks. This is due to the different goals of the two tasks. The single-view association focuses on the temporal continuity of the object embedding without many variants of poses and views; however, cross-view matching focuses on the view-independent features, such as clothes colors, types, and gaits of objects. As a result, we decouple the embedding into cross-view Re-ID and single-view Re-ID heads.

To learn the single-view Re-ID embedding, we follow the similar loss defined in Eq.~(\ref{loss_cross}), \textit{i.e.},
\begin{equation}
    \mathcal{L}_{s} = \sum_{\boldsymbol{I} \in \mathcal{V}}  \phi_s(h_{s}(f(\boldsymbol{I};\boldsymbol{\theta}_f);\boldsymbol{\theta}_{h_s}), \boldsymbol{y}^{LID}),
    \label{loss_single}
\end{equation}
where $\boldsymbol{y}^{LID}$ represents the one-hot vector of the object local ID. The local ID of an object is specific to that video, and objects in different videos have different local IDs even when they have the same global ID.

The cross-entropy loss is a common choice for $\phi_s(\cdot, \cdot)$ in Eq.~(\ref{loss_single}). However, with such a definition, we find there is a large conflict between the cross-view Re-ID loss and single-view Re-ID loss due to the different definitions of ground truth IDs. The same object in different views is treated as positive samples in the cross-view Re-ID, while they are treated as negative samples in the single-view Re-ID, leading to further degradation of the tracking performance from the observations of experimental results. To address this conflict, we define a conflict-free cross-entropy loss as follows,
\begin{equation}
\begin{aligned}
& \phi_s = \\
& -\frac{1}{N_d}\sum_{i=1}^{N_d}\log\frac{e^{\boldsymbol{W}_{y_i}^T\boldsymbol{x}_i+b_{y_i}}}{e^{\boldsymbol{W}_{y_i}^T\boldsymbol{x}_i+b_{y_i}}+\sum_{y_j\neq y_i}\mathbbm{M}_{i,j}e^{\boldsymbol{W}_{y_j}^T\boldsymbol{x}_i+b_{y_j}}},
\label{conflict_free_softmax}
\end{aligned}
\end{equation}
where $\{\boldsymbol{W}_i, b_i\}$ are learnable parameters from the last fully-connected layer of the single-view Re-ID head with respect to the $i$-th local ID; $\boldsymbol{x}_i$ is the input feature to the last layer; $y_i$ is the local ID; 
$N_d$ is the number of objects; and $\mathbbm{M}_{i,j}=\textbf{1}_{v_j=v_i}$ is the indicator function, which returns 1 if only if the local id $y_i$ and $y_j$ are from the same video sequence; otherwise returns 0. In other words, we only compute the softmax cross-entropy loss on objects from the individual video sequence of the same view. Without the cross-view distraction, the single-view Re-ID can address the previous conflict problem. Based on this definition, we use the positive and negative samples in the softmax to remain consistent for both global IDs and local IDs. We demonstrate the process in the bottom-right of Fig.~\ref{fig_fw}.

\subsubsection{Final Loss}
Following \cite{zhang2021fairmot} in FairMOT, we use learnable parameters $w_1$ and $w_2$ to adjust individual losses for training by the uncertainty loss \cite{kendall2018multi}. We formulate the final loss for training CrossMOT as:
\begin{equation}
\mathcal{L}_{total}=\frac{1}{2}(\frac{1}{e^{w_1}}\mathcal{L}_{d}+\frac{1}{e^{w_2}}  (\mathcal{L}_{s}+\mathcal{L}_{c})+w_1+w_2),
\end{equation}
where 
$w_1$ and $w_2$ are learnable parameters used to balance between the detection and Re-ID branches. 

\subsection{Inference of CrossMOT}
\label{sec:inference}
In the inference stage, we first feed the image into the trained network, and the produced detection head is translated into bounding boxes using a decoder. We match each bounding box with corresponding single-view and cross-view features. After that, we choose $n_i$ bounding boxes $\mathcal{B}_i=\{b_i^j\vert j=1,2,...,n_i\}$ in video $\boldsymbol{V}_i$ with confidence greater than detection threshold ${\delta_d}$ and take them into our tracking framework. In the tracking process, single-view and cross-view matching alternate.
We employ DeepSORT \cite{wojke2017simple} for single-view tracking.
For each frame $t$ in video $\boldsymbol{V}_i$, we first calculate the cost matrix $\boldsymbol{C}_i$ using single-view Re-ID embedding and then generate the gate matrix $\boldsymbol{G}_i$ to reduce the excessive value in $\boldsymbol{C}_i$ using single-view matching threshold $\delta_s$. After that, the permutation matrix $\boldsymbol{P}_s^{t,t-1}$ is created by running the Hungarian algorithm \cite{kuhn1955hungarian} to the cost matrix $\boldsymbol{C}_i$.

For cross-view tracking, we follow the MvMHAT \cite{gan2021mvmhat} and calculate the association matrix $\boldsymbol{A}^{ij} \in \mathbbm{R}^{n_i \times n_j}$ of frame $t$ in video $\boldsymbol{V}_i$ and $\boldsymbol{V}_j$ as: $\boldsymbol{A}^{ij} = \boldsymbol{E}_i \cdot \boldsymbol{E}_j^T$, where $\boldsymbol{E}_i \in \mathbbm{R}^{n_i \times K_c}$ and $\boldsymbol{E}_j \in \mathbbm{R}^{n_j \times K_c}$ are cross-view embedding matrices of video $V_i$ and $V_j$, respectively; $K_c$ is the dimension of cross-view embedding. We use a temperature-adaptive softmax operation \cite{hinton2015distilling} to compute the matching matrix $\boldsymbol{M}^{ij}$ as $\boldsymbol{M}^{ij}_{ab}=\frac{exp(\tau \boldsymbol{A}^{ij}_{ab})}{\sum_{b'=1}^{A_c} exp(\tau \boldsymbol{A}^{ij}_{ab'})}$, where $a$, $b$, and $A_c$ denote the row index, column index, and number of columns in $\boldsymbol{A}^{ij}$, respectively; $\tau$, the adaptive temperature, is calculated by two predefined parameters $\epsilon$ and $\gamma$: $\tau=\frac{1}{\epsilon} log[\frac{\gamma(A_c-1)+1}{1-\gamma}]$. All entries in $\boldsymbol{M}^{ij}$ that is less than or equal to the cross-view matching threshold $\delta_c$ are set to 0. Like single-view tracking, we generate the permutation matrix $\boldsymbol{P}^{ij}_c$ by adopting the Hungarian algorithm to $\boldsymbol{M}^{ij}$. 

\section{Experiments}
\label{sec:exp}

\subsection{Experiment Settings and Tasks}
In our dataset, the 75 video sequences are in chronological order, with 25 groups in 15 scenes. 30 videos from the first 10 groups are used for training. We use the rest as testing, where 10 groups are the easy test sets with the same scenes as the train set. We select the other 5 groups without repeated scenes as the hard test set. Our DIVOTrack could be used for the research of object detection, single-view tracking, and cross-view tracking. In this paper, we mainly introduce the settings and tasks of single-view tracking and cross-view tracking.



\begin{table*}[!t]
\small
\caption{Comparison between single-view tracking baseline methods on the easy test set. The best and second-best performances for each column are shown in bold and light blue.}
\centering
\begin{tabular}{l|ccccccccc}
\toprule
Methods & HOTA$\uparrow$   & IDF1$\uparrow$  & MOTA$\uparrow$  & MOTP$\uparrow$  & MT$\uparrow$  & ML$\downarrow$  & AssA$\uparrow$  & IDSw$\downarrow$  & FM$\downarrow$\\
\midrule
Deepsort \cite{wojke2017simple} & 54.3 & 59.9 & \textcolor{cyan}{79.6} & 81.2 & 462 & 50 & 45.0 & 1,920 & \textbf{2,504} \\
CenterTrack \cite{centertrack} & 55.3 & 62.2 & 73.4 & 80.6 & \textbf{534} & \textcolor{cyan}{35} & 49.2 & 1,631 & 2,950  \\
Tracktor \cite{tracktor_2019_ICCV} & 48.4 & 56.2 & 66.6 & 80.8 & \textcolor{cyan}{517} & \textbf{22} & 40.3 & 1,382 & 3,337 \\
FairMOT \cite{zhang2021fairmot} & \textbf{65.3} & \textbf{78.2} & \textbf{82.7} & \textcolor{cyan}{81.9} & 486 & 48 & \textbf{62.7} & \textbf{731} & 3,498\\
TraDes \cite{Wu2021TraDeS} & \textcolor{cyan}{58.9} & \textcolor{cyan}{67.3} & 74.2 & \textbf{82.3} & 504 & 38 & \textcolor{cyan}{54.0} & \textcolor{cyan}{1,263} & \textcolor{cyan}{2,647}\\
\bottomrule
\end{tabular}
\label{tab:singelview_easy}
\end{table*}

\begin{table*}[!t]
\small
\caption{Comparison between single-view tracking baseline methods on the hard test set. The best and second-best performances for each column are shown in bold and light blue.}
\centering
\begin{tabular}{l|ccccccccc}
\toprule
Methods & HOTA$\uparrow$   & IDF1$\uparrow$  & MOTA$\uparrow$  & MOTP$\uparrow$  & MT$\uparrow$  & ML$\downarrow$  & AssA$\uparrow$  & IDSw$\downarrow$  & FM$\downarrow$\\
\midrule
Deepsort \cite{wojke2017simple} & 44.0 & 44.7 & \textcolor{cyan}{62.6} & \textcolor{cyan}{82.5} & \textcolor{cyan}{325} & \textcolor{cyan}{173} & 34.1 & 2,547 & 3,412 \\
CenterTrack \cite{centertrack} & 44.1 & 46.3 & 55.9 & 81.1 & 320 & 221 & 37.2 & 2,244& \textbf{2,954}  \\
Tracktor \cite{tracktor_2019_ICCV} & 38.4 & 43.2 & 53.3 & 80.7 & \textbf{357} & \textbf{137} & 29.0 & 1,903 & 4,405 \\
FairMOT \cite{zhang2021fairmot} & \textbf{56.5} & \textbf{64.3} & \textbf{66.7} & \textbf{84.3} & 270 & 222 & \textbf{54.7} & \textbf{954} & 5,614\\
TraDes \cite{Wu2021TraDeS} & \textcolor{cyan}{46.1} & \textcolor{cyan}{52.2} & 61.8 & 79.9 & 311 & 234 & \textcolor{cyan}{39.6} & \textcolor{cyan}{1,656} & \textcolor{cyan}{3,295}\\
\bottomrule
\end{tabular}
\label{tab:singelview_hard}
\end{table*}

\subsubsection{Single-View Tracking Setting}

We treat each video sequence independently for single-view trackers. 
We employ ID F1 measure (IDF1) \cite{ristani2016performance}, higher order tracking accuracy (HOTA) \cite{luiten2020hota}, multiple object tracking accuracy (MOTA), multiple object tracking precision (MOTP), mostly tracked targets (MT), mostly lost targets (ML), association accuracy (AssA), fragments (FM) and identity switches (IDSw) \cite{milan2016mot16} as the evaluation metrics of the tracking performance, which are widely used in MOT.

\subsubsection{Cross-View Tracking Setting}
Cross-view trackers, unlike single-view trackers, process multiple views within each batch of synchronized videos.
If the same object appears in different recordings from the same group, the object should have the same ID.

As for evaluation, we use the metrics introduced in \cite{han2020cvmht} as the standardized measurements, in which cross-view ID F1 metric (CVIDF1) and cross-view matching accuracy (CVMA) are proposed based on IDF1 and MOTA metrics. Specifically, CVIDF1 and CVMA are defined as follows,
\begin{equation}
    \textrm{CVIDF1} = \frac{ 2\textrm{CVIDP} \times \textrm{CVIDR}}{\textrm{CVIDP}+\textrm{CVIDR}},
\end{equation}
\begin{equation}
    \textrm{CVMA} = 1-(\frac{\sum_t \textrm{m}_t+\textrm{fp}_t+2\textrm{mme}_t}{\sum_t \textrm{gt}_t}),
\label{eq:cvma}
\end{equation}
where CVIDP and CVIDR denote the cross-view object matching precision and recall, respectively. $\textrm{m}_t$, $\textrm{fp}_t$, $\textrm{mme}_t$, and $\textrm{gt}_t$ are the numbers of misses, false positives, mismatched pairs, and the total number of objects in all views at time $t$, respectively.

\begin{table*}[!t]
\small
\centering
\caption{Comparison between cross-view tracking baseline methods on DIVOTrack with CVMA (``CA'') and CVIDF1 (``C1''). The best and second-best performances for each column are shown in bold and light blue. The first ten scenes are the easy test set, and the last five scenes are the hard test set.}\label{tab:crossview}
\begin{tabular}{L{2.2cm}|C{0.8cm}C{0.8cm}|C{0.8cm}C{0.8cm}|C{0.8cm}C{0.8cm}|C{0.8cm}C{0.8cm}|C{0.8cm}C{0.8cm}}
\toprule
Scenes & \multicolumn{2}{|c}{Circle} & \multicolumn{2}{|c}{Shop} & \multicolumn{2}{|c}{Moving} & \multicolumn{2}{|c}{Park} & \multicolumn{2}{|c}{Ground} \\
\midrule
Methods & CA$\uparrow$ & C1$\uparrow$  & CA$\uparrow$  & C1$\uparrow$  & CA$\uparrow$  & C1$\uparrow$ & CA$\uparrow$  & C1$\uparrow$ & CA$\uparrow$  & C1$\uparrow$ \\
\cmidrule(r){1-11}
OSNet \cite{zhou2019osnet} & 36.0 & 47.2 & 53.4 & 50.1 & 29.6 & 50.1 & 30.7 & 43.3 & 23.6 & 38.6\\
Strong \cite{Luo_2019_Strong_TMM} & 37.3 & 43.9 & 56.6 & 42.1 & 37.0 & 45.4 & 38.4 & 48.5 & 29.9 & 38.2\\
AGW \cite{pami21reidsurvey} & 56.9 & 56.9 & 60.7 & 47.1 & 35.3 & 47.3 & 55.1 & 63.6 & 50.6 & 50.3\\
MvMHAT \cite{gan2021mvmhat} & 69.2 & 66.3 & \textcolor{cyan}{62.6} & \textcolor{cyan}{53.4} & \textcolor{cyan}{47.6} & \textbf{58.5} & 57.9 & 64.9 & 50.2 & 51.1\\
CT \cite{wieczorek2021unreasonable} & \textcolor{cyan}{71.2} & \textcolor{cyan}{68.3} & \textcolor{cyan}{62.6} & 52.8 & 42.4 & \textcolor{cyan}{56.2} & \textcolor{cyan}{65.4} & \textcolor{cyan}{71.4} & \textcolor{cyan}{58.8} & \textcolor{cyan}{55.9}\\
MGN \cite{wang2018learning} & 36.1 & 42.5 & 52.4 & 40.2 & 27.1 & 36.9 & 30.7 & 41.9 & 24.2 & 31.1\\
\textbf{CrossMOT} & \textbf{74.0} & \textbf{74.2} & \textbf{66.4} & \textbf{54.8} & \textbf{58.3} & 53.0 & \textbf{78.9} & \textbf{80.5} & \textbf{65.8} & \textbf{66.2} \\
\midrule

Scenes  & \multicolumn{2}{|c}{Gate1} & \multicolumn{2}{|c}{Floor} & \multicolumn{2}{|c}{Side} & \multicolumn{2}{|c}{Square} & \multicolumn{2}{|c}{Gate2} \\
\midrule
Methods & CA$\uparrow$ & C1$\uparrow$  & CA$\uparrow$  & C1$\uparrow$  & CA$\uparrow$  & C1$\uparrow$ & CA$\uparrow$  & C1$\uparrow$ & CA$\uparrow$  & C1$\uparrow$ \\
\cmidrule(r){1-11}
OSNet \cite{zhou2019osnet} & 26.0 & 47.7 & 36.8 & 42.9 & 48.9 & 54.9 & 24.9 & 42.5 & 23.6 & 48.9\\
Strong \cite{Luo_2019_Strong_TMM} & 39.7 & 55.7 & 44.4 & 38.8 & 54.6 & 56.6 & 36.0 & 46.1 & 26.7 & 51.9\\
AGW \cite{pami21reidsurvey} & 58.2 & 67.6 & 60.9 & 49.1 & 63.4 & 61.9 & 49.9 & 56.6 & 81.7 & \textcolor{cyan}{89.0}\\
MvMHAT \cite{gan2021mvmhat} & 57.8 & 70.1 & 65.6 & \textbf{62.3} & 67.0 & 68.8 & 58.8 & 69.2 & \textcolor{cyan}{91.3} & 88.4\\
CT \cite{wieczorek2021unreasonable} & \textcolor{cyan}{66.3} & \textcolor{cyan}{76.3} & \textcolor{cyan}{67.1} & \textcolor{cyan}{53.8} & \textcolor{cyan}{70.2} & \textcolor{cyan}{71.8} & \textcolor{cyan}{59.7} & \textcolor{cyan}{71.5} & 90.2 & \textbf{94.9}\\
MGN \cite{wang2018learning} & 24.7 & 42.1 & 33.8 & 32.3 & 45.7 & 47.0 & 25.4 & 41.0 & 22.9 & 47.9\\
\textbf{CrossMOT} & \textbf{78.2} & \textbf{82.3} & \textbf{76.7} & \textbf{62.3} & \textbf{78.8} & \textbf{81.3} & \textbf{67.2} & \textbf{75.9} & \textbf{91.5} & 87.1 \\
\midrule

Scenes & \multicolumn{2}{|c}{Indoor1} & \multicolumn{2}{|c}{Indoor2} & \multicolumn{2}{|c}{Outdoor1} & \multicolumn{2}{|c}{Outdoor2} & \multicolumn{2}{|c}{Park2} \\
\midrule
Methods & CA$\uparrow$ & C1$\uparrow$  & CA$\uparrow$  & C1$\uparrow$  & CA$\uparrow$  & C1$\uparrow$ & CA$\uparrow$  & C1$\uparrow$ & CA$\uparrow$  & C1$\uparrow$ \\
\cmidrule(r){1-11}
OSNet \cite{zhou2019osnet} & 43.1 & 43.6 & 35.8 & 39.7 & 37.7 & 32.3 & 9.0 & 31.0 & 30.2 & 46.6\\
Strong \cite{Luo_2019_Strong_TMM} & 44.4 & 39.6 & 36.7 & 41.1 & 42.3 & 35.8 & 12.6 & 28.3 & 32.6 & 38.3\\
AGW \cite{pami21reidsurvey} & 44.4 & 39.6 & 36.9 & 40.4 & 53.0 & 47.3 & 18.7 & 35.1 & 38.0 & 42.1\\
MvMHAT \cite{gan2021mvmhat} & \textcolor{cyan}{47.0} & \textcolor{cyan}{50.1} & 36.5 & \textcolor{cyan}{42.8} & \textcolor{cyan}{61.5} & 55.7 & \textcolor{cyan}{31.0} & \textbf{54.6} & \textbf{49.2} & \textbf{64.5}\\
CT \cite{wieczorek2021unreasonable} & 46.3 & 45.4 & \textcolor{cyan}{37.2} & 42.0 & 59.1 & 56.1 & 21.3 & 41.8 & 45.0 & 52.3\\
MGN \cite{wang2018learning} & 44.7 & 42.2 & \textcolor{cyan}{37.2} & 42.5 & \textbf{61.6} & \textcolor{cyan}{56.7} & 28.2 & 41.8 & 48.5 & 50.7\\
\textbf{CrossMOT} & \textbf{56.1} & \textbf{54.3} & \textbf{52.0} & \textbf{60.3} & 61.1 & \textbf{57.3} & \textbf{36.1} & \textcolor{cyan}{54.0} & \textcolor{cyan}{49.0} & \textcolor{cyan}{56.0} \\
\bottomrule
\end{tabular}
\end{table*}

\subsection{Implementation Details of CrossMOT}
We adopt DLA-34 \cite{duan2019centernet} as our backbone network. We use the pre-trained model on COCO dataset \cite{lin2014microsoft} to initialize our model. In our backbone, the resolution of the feature map is 272$\times$152, and the size of the input image is resized to four times as the feature map, \textit{i.e.}, 1088$\times$608. The feature dimension of cross-view embedding and single-view embedding are both set to 512. We train our model with the Adam optimizer \cite{kingma2014adam} for 30 epochs with a start learning rate $10^{-4}$ and batch size of 8. The learning rate decays to $10^{-5}$ at 20 epochs. The loss function balance parameters $w_1$ and $w_2$ are set to $-1.85$ and $-1.05$ at the start of training, following \cite{zhang2021fairmot}. We set the detection threshold, single-view distance threshold, and cross-view matching threshold to $\delta_d = 0.5$, $\delta_s=0.3$ and $\delta_c=0.5$, respectively. We use two parameters for calculating the adaptive temperature $\tau$ are set to $\epsilon=0.5$ and $\gamma=0.5$, following \cite{hinton2015distilling}. We train/test our model on a single NVIDIA RTX 3090 24GB GPU.

\subsection{Single-View Tracking Baselines on DIVOTrack}


We compare five widely used single-view tracking methods, including Deepsort \cite{wojke2017simple}, CenterTrack \cite{centertrack}, Tracktor \cite{tracktor_2019_ICCV}, FairMOT \cite{zhang2021fairmot}, and TraDes \cite{Wu2021TraDeS}. 


We use the default configurations for training and testing the aforementioned trackers. Note that we finetune the detector of Tracktor with 5 epochs and finetune TraDes with 30 epochs based on its pre-trained model. We train all the models using four Nvidia RTX 3090 GPUs. 
We provide the comparison of baseline methods in Table~\ref{tab:singelview_easy} and Table~\ref{tab:singelview_hard}. FairMOT is a strong baseline single-view MOT method and performs better than other trackers on the easy test set, where the HOTA, IDF1, and MOTA of FairMOT are $65.3\%$, $78.2\%$, and $82.7\%$, respectively. On the hard test set, FairMOT also has better results on HOTA, IDF1, and MOTA than other trackers. The results prove that a feature embedding network can significantly improve tracking performance with detections. 
We can see that there are large variations in performances from different methods. For example, HOTA ranges from $48.4\%$ to $65.3\%$ on the easy test set and ranges from $38.4\%$ to $56.5\%$ on the hard test set, demonstrating DIVOTrack dataset has discrimination for different trackers.

\subsection{Cross-View Tracking}
\label{sec:crossview}

To evaluate the performance of cross-view tracking, we first obtain the detection results from the trained CenterNet model, then follow different embedding networks \cite{zhou2019osnet,Luo_2019_Strong_TMM,pami21reidsurvey,gan2021mvmhat} to obtain the object features, and finally achieve the cross-view tracking following the association framework \cite{gan2021mvmhat}. We adopt six different feature embedding networks for the comparison, including OSNet \cite{zhou2019osnet}, Strong \cite{Luo_2019_Strong_TMM}, AGW \cite{pami21reidsurvey}, MvMHAT \cite{gan2021mvmhat}, Centroids (CT) \cite{wieczorek2021unreasonable}, MGN \cite{wang2018learning}, and our proposed CrossMOT method. 

\subsubsection{Cross-View Tracking Results on DIVOTrack}

We report the detailed results of all baseline methods on each scene of DIVOTrack in Table~\ref{tab:crossview}.
Our proposed method CrossMOT outperforms other cross-view tracking methods of almost all the fifteen scenarios, showing the effectiveness of CrossMOT. 
For the easy test set, the \textit{Ground} scene has worse performance than other scenes of all the methods. Due to the \textit{Indoor2} scene contains more cross-view pedestrians, as seen in Fig.~\ref{fig:cv_crossbox}. The dense cross-view objects provide additional difficulties for the trackers. As for \textit{Gate2}, tracking becomes significantly easier since there are fewer objects in the scene. Our CrossMOT outperforms other methods in this scene and demonstrates its generalization to diverse scenes. The detailed results demonstrate the diversity of different scenes proposed in DIVOTrack.



\begin{table*}[!t]
\small
\centering
\caption{Cross-view tracking results on DIVOTrack and other existing datasets with CVMA (``CA'') and CVIDF1 (``C1''). ``DIVO.E'' is the DIVOTrack easy test set and ``DIVO.H'' is the hard test set. The best and second-best performances for each column are shown in bold and light blue.}\label{tab:existdataset}
\begin{tabular}{l|cc|cc|cc|cc|>{\columncolor{Gray}}c>{\columncolor{Gray}}c>{\columncolor{Gray}}c>{\columncolor{Gray}}c}
\toprule
~ & \multicolumn{2}{|c|}{EPFL} & \multicolumn{2}{|c|}{CAMPUS} & \multicolumn{2}{|c|}{MvMHAT}& \multicolumn{2}{|c}{WILDTRACK}& \multicolumn{2}{|>{\columncolor{Gray}}c}{\textbf{DIVO.E}} & \multicolumn{2}{|>{\columncolor{Gray}}c}{\textbf{DIVO.H}} \\
\midrule
Methods & CA$\uparrow$ & C1$\uparrow$  & CA$\uparrow$ & C1$\uparrow$ & CA$\uparrow$ & C1$\uparrow$ & CA$\uparrow$ & C1$\uparrow$ & CA$\uparrow$ & C1$\uparrow$ & CA$\uparrow$ & C1$\uparrow$ \\
\cmidrule(r){1-13}
OSNet \cite{zhou2019osnet} & 73.0 & 40.3 & 58.8 & 47.8 & \textbf{92.6} & \textbf{87.7} & 10.8 & 18.2 & 34.3 & 46.0 & 30.7 & 38.0 \\
Strong \cite{Luo_2019_Strong_TMM} & \textbf{75.6} & \textcolor{cyan}{45.2} & 63.4 & 55.0 & 49.0 & 55.1 & 28.6 & 41.6 & 40.9 & 45.8 & 33.0 & 36.4\\
AGW \cite{pami21reidsurvey} & 73.9 & 43.2 & 60.8 & 52.8 & \textcolor{cyan}{92.5} & 86.6 & 15.6 & 23.8 & 57.0 & 56.8 & 36.6 & 40.0\\
MvMHAT \cite{gan2021mvmhat} & 30.5 & 33.7 & 56.0 & 55.6 & 70.1 & 68.4 & 10.3 & 16.2 & 61.1 & 62.6 & \textcolor{cyan}{42.6} & \textcolor{cyan}{51.5}\\
CT \cite{wieczorek2021unreasonable} & 75.5 & 45.1 & \textcolor{cyan}{63.7} & 55.0 & 46.7 & 53.5 & 19.0 & 42.0 & \textcolor{cyan}{64.9} & \textcolor{cyan}{65.0} & 39.4 & 45.7 \\
MGN \cite{wang2018learning} & 73.3 & 42.6 & 63.3 & \textcolor{cyan}{56.1} & 
92.3 & \textcolor{cyan}{87.4} & \textcolor{cyan}{32.6} & \textcolor{cyan}{46.2} & 33.5 & 39.4 & 41.4 & 45.0\\
\textbf{CrossMOT} & \textcolor{cyan}{74.4} & \textbf{47.3} & \textbf{65.6} & \textbf{61.2} & 92.3 & \textcolor{cyan}{87.4} & \textbf{42.3} & \textbf{56.7} & \textbf{72.4} & \textbf{71.1}& \textbf{50.0} & \textbf{56.3}\\
\bottomrule
\end{tabular}
\end{table*}

\subsubsection{Cross-View Tracking Results on Other Datasets}

We provide the results of these approaches on DIVOTrack and other aforementioned existing datasets in Table~\ref{tab:existdataset}. 
On the other datasets like CAMPUS and WILDTRACK, CrossMOT also outperforms other methods, which proves the efficacy of our proposed approach. For the MvMHAT dataset, OSNet and AGW achieve relatively good performance. This is because videos from MvMHAT are collected from the same scenario and share identical subjects. On EPFL, CrossMOT also outperforms other methods, showing that CrossMOT is more suitable for complex real-world scenarios. On the hard test set our CrossMOT also outperforms other methods, which demonstrates its generalization ability to unseen datasets.

\begin{figure}[!t]
\begin{center}
\includegraphics[width=1\linewidth]{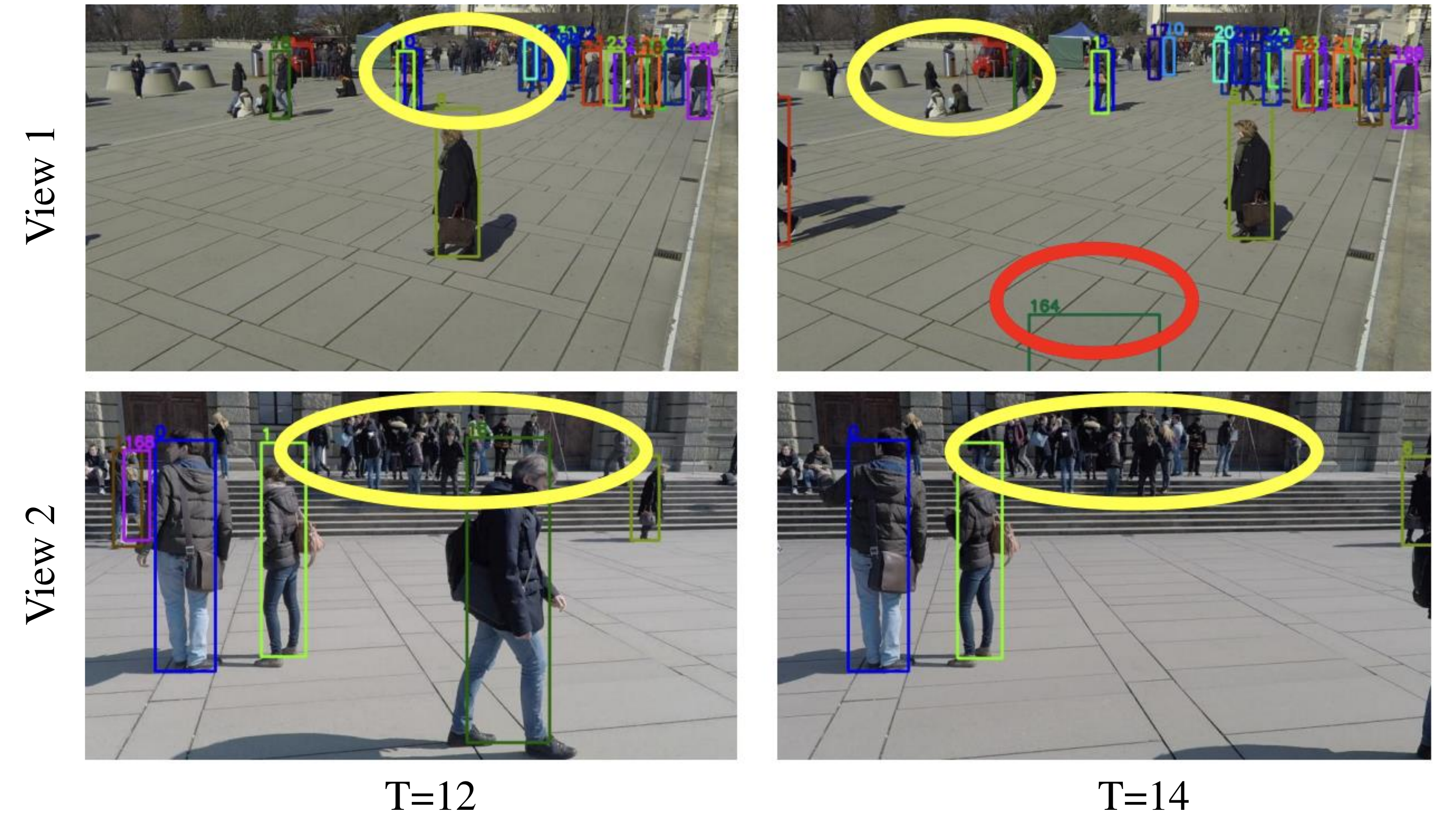}
\end{center}
   \caption{Some examples of noisy annotations on WILDTRACK. Yellow circles represent the missing annotations, and red circles represent incorrect boxes. ``T'' is the frame index.}
\label{fig:quality_wild}
\end{figure}

As shown in Table~\ref{tab:existdataset}, most of the methods have better results on EPFL, CAMPUS, and MvMHAT datasets because of limited scenes and the limited number of subjects. In addition, WILDTRACK only has 1 scene and missing lots of annotations, which causes worse results. The noisy annotations on WILDTRACK are shown in Fig.~\ref{fig:quality_wild}. Comparing other datasets, DIVOTrack has more diverse scenes and a large number of annotated subjects. Moreover, the metrics of all methods have a significant decline on the hard test set, proving that they still have room for improvement on DIVOTrack.

\subsubsection{Qualitative Results of CrossMOT} 
\begin{figure*}[htb]
\centering
\includegraphics[width=1\textwidth]{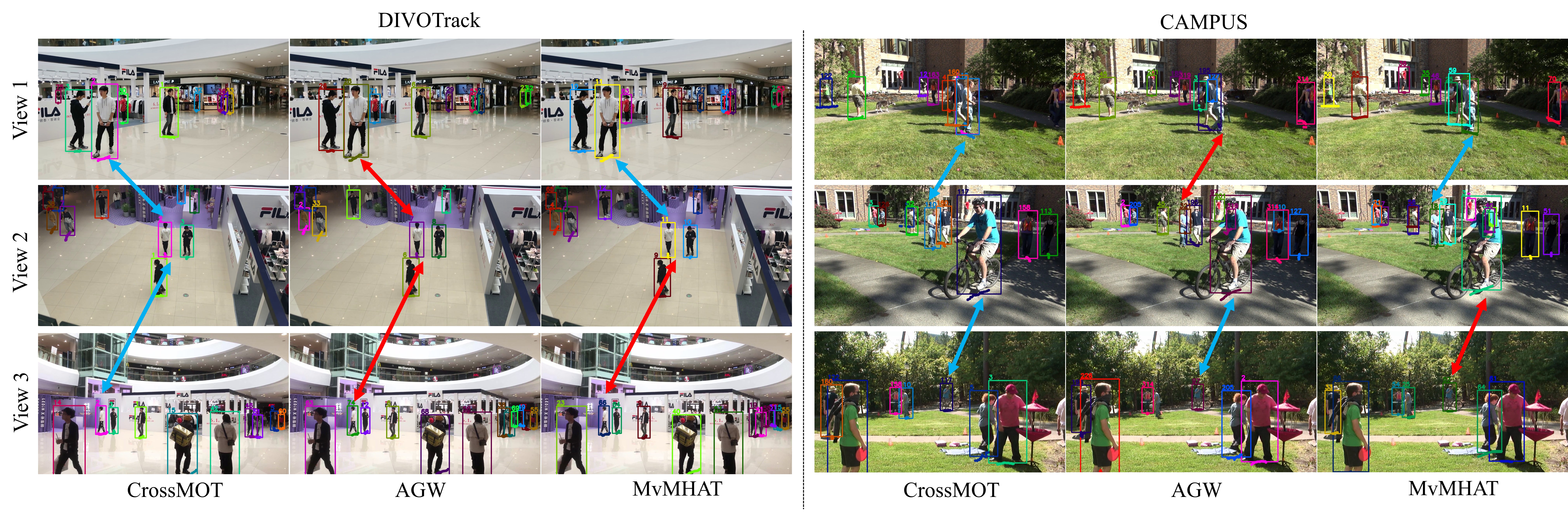} 
\caption{Qualitative examples of cross-view tracking performance for the proposed CrossMOT, AGW, and MvMHAT on DIVOTrack and CAMPUS datasets. Rows and columns represent camera views and different methods, respectively. Blue and red arrows represent correctly matched pairs and mis-matched pairs, respectively.}
\label{fig_ex}
\end{figure*}

To better show the effectiveness of our CrossMOT, we show some qualitative examples in Fig.~\ref{fig_ex}. Left and right sub-figures are results on DIVOTrack and CAMPUS datasets, respectively. Rows represent camera views, and columns represent different methods. Blue and red arrows represent correctly matched pairs and mis-matched pairs, respectively. Compared with other baseline methods, CrossMOT generates much fewer cross-view matching errors, demonstrating the effectiveness of our method.

\subsection{Ablation Studies of CrossMOT} 

\subsubsection{Different Variants of the Model}

We show a comparison between three variants of the proposed model on DIVOTrack easy test set, MvMHAT and CAMPUS datasets in Table~\ref{tab:variant}. The three variants include \textit{shared emb.}, \textit{w/o conflict-free}, and \textit{full model}. Specifically, \textit{Shared emb.} represents using a single Re-ID head for both single-view and cross-view embedding, and \textit{w/o conflict-free} represents using the original cross-entropy loss without conflict-free loss for embedding. For DIVOTrack and MvMHAT, we see consistent improvements in the full model compared with the other two variants. On CAMPUS, our full model outperforms other methods on CVMA and is close to the best result on CVIDF1. These results verify the effectiveness of our decoupled multi-head embedding strategy and the designed conflict-free loss.

\subsubsection{Variants on the Tracking Inference Thresholds} 
\begin{figure}[!t]
\centering
\includegraphics[width=1\linewidth]{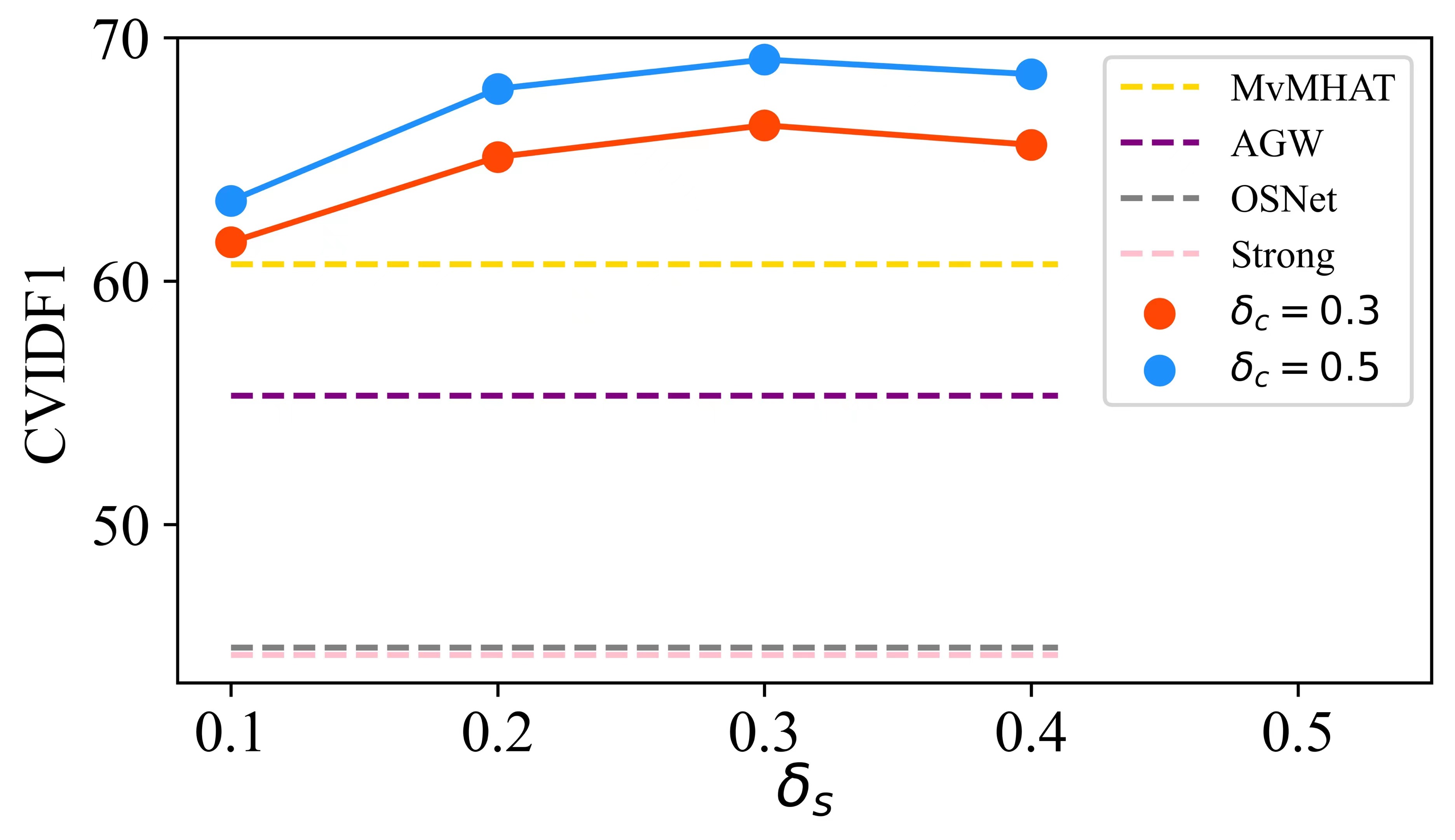} 
\caption{The CVIDF1 of the proposed method (with solid line) with variant thresholds on the DIVOTrack easy test set. Red and blue curves represent the proposed method with $\delta_c=0.3$ and $\delta_c=0.5$, respectively. The x-axis represents the change of $\delta_s$. The performances of the other four baseline methods (with dashed lines) are shown for reference.}
\label{fig_thres}
\end{figure}

We vary the tracking thresholds, \textit{i.e.}, $\delta_c$, and $\delta_s$, and see the influence on the final results. And we conduct experiments on the DIVOTrack easy test set. We vary $\delta_s$ from 0.1 to 0.4 when $\delta_c \in \{0.3,0.5\}$. We also provide the other four baseline methods for reference in Fig.~\ref{fig_thres}. 
Although there are some fluctuations in the results for selecting different thresholds, our proposed method can consistently outperform other methods, demonstrating the robustness of our proposed method. 

\begin{table}[!t]
\centering
\caption{Comparison between different variants of the proposed model on DIVOTrack, MvMHAT, and CAMPUS datasets. \textit{Shared emb.} represents using a single Re-ID head for both single-view and cross-view embedding. \textit{W/O conflict-free} represents using the original cross-entropy loss without the conflict-free loss for embedding. The best and second-best performances for each column are shown in bold and light blue.}
\label{tab:variant}
\begin{tabular}{L{1.4cm}|L{2.4cm}|C{0.8cm}C{0.8cm}}

\toprule
Dataset & Variant & CVMA & CVIDF1\\
\midrule
\multirow{3}*{DIVOTrack} &
Shared Emb. & \textcolor{cyan}{70.7} & \textcolor{cyan}{70.9} \\
& W/O Conflict-free & 70.1 & 69.8\\
& \textbf{Full Model} & \textbf{72.4} & \textbf{71.1} \\
\midrule

\multirow{3}*{MvMHAT} &
Shared Emb. & \textcolor{cyan}{92.2} & 86.2 \\
& W/O Conflict-free & 91.9 & \textcolor{cyan}{87.0}\\
& \textbf{Full Model} & \textbf{92.3} & \textbf{87.4}\\

\midrule
\multirow{3}*{CAMPUS} &
Shared Emb. & \textcolor{cyan}{64.8} & \textbf{61.9} \\
& W/O Conflict-free & 64.4 & 58.0 \\
& \textbf{Full Model} & \textbf{65.6} & \textcolor{cyan}{61.2}\\

\bottomrule
\end{tabular}

\end{table}

\subsection{Benefits from DIVOTrack}

Our DIVOTrack benchmark has several benefits compared with existing benchmarks \cite{fleuret2007multicamera,xu2017cross,gan2021mvmhat,chavdarova2018wildtrack}. 
\textbf{First}, one publicly accessible detector is used for the baseline methods if they follow the tracking-by-detection framework. However, the publicly accessible detection results are missing in existing benchmarks \cite{xu2017cross,gan2021mvmhat}. This may cause unfair comparisons for methods that are applied in such benchmarks. 
\textbf{Second}, some of the existing benchmarks do not provide clear cross-view tracking results, where only single-view tracking metrics are used in the evaluation \cite{fleuret2007multicamera,xu2017cross,chavdarova2018wildtrack}. 
\textbf{Third}, the detailed performance of each scene is analyzed, where the influence of different backgrounds of environments on the tracking performance is demonstrated. Other benchmarks do not support this since most of the previous datasets do not contain diverse scenes.
\textbf{In addition}, in our benchmark, we also release a unified framework that can combine the Re-ID embedding networks \cite{zhou2019osnet,Luo_2019_Strong_TMM,pami21reidsurvey} in the cross-view tracking. Researchers are free to verify the performance of applying any open-sourced Re-ID embedding networks in our cross-view tracking.
\textbf{Last}, we also provide the source code of all the compared baseline methods. We summarize the benefits of our benchmark in Table~\ref{tab:datacom}. Our benchmark provides accessible public detections (Det), cross-view evaluations (CE), individual scene-based analysis (SA), accessible cross-view framework (CF), and cross-view baseline methods (CVBM).

\begin{table}[!t]
\small
\centering
\caption{Comparison of benchmark evaluations, including EPFL \cite{fleuret2007multicamera}, CAMPUS ("CAM.") \cite{xu2017cross}, MvMHAT ("Mv.") \cite{gan2021mvmhat}, WILDTRACK ("WILD.") \cite{chavdarova2018wildtrack} and DIVOTrack ("DIVO.").}
\label{tab:datacom}
\begin{tabular}{c|ccccc}
\toprule
Benchmarks & Det & CE & SA & CF & CVBM \\
\midrule
WILD. & - & - & - & - & -\\
EPFL & - & \checkmark & \checkmark & - & -\\
CAM. & - & - & \checkmark & - & - \\
Mv. & - & \checkmark & - & \checkmark & \checkmark\\
\rowcolor{Gray}
\textbf{DIVO.} (Ours) & \checkmark & \checkmark & \checkmark & \checkmark & \checkmark\\
\bottomrule
\end{tabular}
\end{table}



\section{Ethical Concerns and Broader Impacts}
\label{sec:ethical}
Because the dataset includes human subjects, some people may have concerns about privacy issues. 
In our dataset, most of the pedestrians in the videos are far away from cameras, with indistinguishable human IDs. For the people who are close to the camera, we will add mosaics on the human faces to protect the privacy of the pedestrians before the dataset is released.
Besides, cross-view tracking is beneficial for smart monitoring and multi-agent-based perception and intelligence. The dataset provides an essential setting and evaluation for such real-world applications.

\section{Conclusion and Future Work}
\label{sec:conclude}

In this paper, we propose a novel cross-view multi-object tracking dataset, namely \textit{DIVOTrack}, which is more realistic, has more tracks and diverse environments and incorporates moving cameras. Accordingly, we build a standardized benchmark for cross-view tracking, with a clear split of training and test sets, publicly accessible detection, and standard cross-view tracking evaluation metrics. We also propose a novel end-to-end cross-view tracking baseline, CrossMOT, which integrates object detection, single-view tracking, and cross-view tracking in a unified embedding model. Our CrossMOT adopts decoupled multi-head embedding that simultaneously learns object detection, single-view Re-ID and cross-view Re-ID. Moreover, we design a locality-aware and conflict-free loss function for single-view embedding to address the ID conflict issue between cross-view embedding and single-view embedding.
With the proposed dataset, benchmark and baseline, the cross-view tracking methods can be fairly compared in the future, which will improve the development of cross-view tracking techniques. 

In future work, we will continuously collect more videos in different weather conditions to enlarge the dataset since the weather condition is not analyzed in the current work. And we will consider improve the annotation quality by segmentation tasks \cite{voigtlaender2019mots, athar2023burst}. For cross-view tracking methods, there are still unresolved issues. We will also try to design an end-to-end joint detection and tracking framework that can take multiple views with variant spatial-temporal relations. In addition, we will also explore how to better utilize the cross-frame and cross-view geometry consistency.

\section*{Acknowledgments}
The authors would also like to thank Tianqi Liu, Zining Ge, Kuangji Chen, Xubin Qiu, Shitian Yang, Jiahao Wei, Yuhao Ge, Hao Chen, Bingqi Yang, Kaixun Jin, Zeduo Yu and Donglin Gu for their work on the dataset collection and annotation.
This work is supported by National Natural Science Foundation of China (62106219).


\bibliography{ref}
\bibliographystyle{abbrv}


\end{document}